\documentclass[runningheads]{llncs}
\usepackage{algorithm}
\usepackage{algpseudocode}
\usepackage[T1]{fontenc}
\usepackage{graphicx}
\usepackage{esvect}
\usepackage{amsmath,amssymb,bm}
\usepackage{amsmath}
\usepackage{amsfonts}
\usepackage{multirow}
\usepackage{enumitem}
\usepackage{graphicx}
\usepackage{textcomp}
\usepackage{booktabs}
\usepackage{mathtools}
\usepackage{microtype}
\usepackage[colorlinks, bookmarksopen, bookmarksnumbered, citecolor=blue, linkcolor=blue, urlcolor=blue]{hyperref}
\let\subparagraph\relax
\usepackage[capitalize]{cleveref}
\usepackage{placeins} % To control float placement
\usepackage{titlesec}
\titlespacing*{\section}{0pt}{0.4\baselineskip}{0.2\baselineskip}
\titlespacing*{\subsection}{0pt}{0.4\baselineskip}{0.2\baselineskip}
\titlespacing*{\subsubsection}{0pt}{0.9\baselineskip}{0.9\baselineskip}
\begin{document}
%
%\title{\texorpdfstring{An Edge--Fog--Cloud Architecture and its Reinforcement Learning Scheduling for the Medical Internet of Things\thanks{This work was supported by Hong Kong Metropolitan University.}}{An Edge-Fog-Cloud Architecture and its Reinforcement Learning Scheduling for the Medical Internet of Things}}

\title{Cloud-Fog-Edge Collaborative Computing For Sequential MIoT Workflow: A Two-tier DDPG-based Scheduling Framework}

\titlerunning{Two-tier DDPG Scheduling for the Medical Internet of Things}

% Second Author\inst{2}\orcidID{1111-2222-3333-4444}}
% \authorrunning{F. Author and S. Author}
% \institute{Affiliation One, City, Country \\
% \email{first.author@example.edu}\\
% Affiliation Two, City, Country \\
% \email{second.author@example.edu}}
%
\author{Yuhao Fu\inst{1}\orcidID{0009-0000-4483-518X}
\and 
Yinghao Zhang\inst{2}\orcidID{0009-0008-2020-869X}
\and
Yalin Liu\inst{1,*}\orcidID{0000-0003-2870-4598}
\and
Bishenghui Tao\inst{1}\orcidID{0000-0003-0968-2346}
\and 
Junhong Ruan\inst{3}\orcidID{0009-0001-9316-6688}
}
\institute{Hong Kong Metropolitan University, Hong Kong, China\\ 
\and  Guangdong Key Lab of AI and Multi-Modal Data Processing, Beijing Normal-Hong Kong Baptist University
\and Hong Kong University of Science and Technology, Hong Kong, China
\email{{ylliu}@hkmu.edu.hk}
}

\maketitle              % typeset the header of the contribution
\begin{abstract}

The Medical Internet of Things (MIoT) demands stringent end-to-end latency guarantees for sequential healthcare workflows deployed over heterogeneous cloud--fog--edge infrastructures. Scheduling these sequential workflows to minimize makespan is an NP-hard problem. To tackle this challenge, we propose a Two-tier DDPG-based scheduling framework that decomposes the scheduling decision into a hierarchical process: a global controller performs layer selection (edge, fog, or cloud), while specialized local controllers handle node assignment within the chosen layer. The primary optimization objective is the minimization of the workflow makespan. Experiments results validate our approach, demonstrating increasingly superior performance over baselines as workflow complexity rises. This trend highlights the framework's ability to learn effective long-term strategies, which is critical for complex, large-scale MIoT scheduling scenarios.

\keywords{MIoT \and Cloud–Fog–Edge Computing \and Sequencial Workflow Scheduling \and Two-tier Reinforcement Learning \and DDPG \and Makespan Minimization.}
\end{abstract}
%
%

%Introduction 的核心包装思路是将全文内容浓缩并逻辑化地呈现。我们首先统一了 sequential MIoT workflow 和 makespan 等核心术语，确保与第二章的系统模型，问题定义对接；接着，将关键的第三段重塑为两个与后续章节强绑定的技术挑战：一是问题本身的 NP-hard 复杂性（第二章optimization problem论证），其核心在于导航第二章定义的“计算-通信权衡”，二是标准“扁平化”RL方法在层级化架构中因动作空间巨大而导致的扩展性瓶颈。这两个挑战直接论证了我们第三章提出的“两层式”DDPG框架的必要性，即通过global控制器进行global权衡、local控制器解决扩展性问题，它提出的问题由后续章节精确解答，从而确保了整篇论文从引言到结论的逻辑连贯性和高度一致性。

\section{Introduction}
The Medical Internet of Things (MIoT), driven by advancements in sensor networks and communication, plays a critical role in supporting smart healthcare applications like real-time patient monitoring and medical image analysis~\cite{Garey1979}. A complete healthcare procedure often comprises a complex \textit{sequential MIoT workflow}, where tasks with diverse computational, memory, and temporal requirements must be executed in a specific order~\cite{Shao25}. While traditional cloud computing offers powerful centralized resources, its reliance on remote data centers introduces significant network latency, making it ill-suited for time-sensitive medical applications~\cite{Alatoun22}. To overcome this, a \textit{cloud-fog-edge collaborative computing} architecture has emerged as a promising solution. By distributing resources, it offers the flexibility to minimize the total workflow execution time, or \textbf{makespan}, by strategically placing tasks closer to the data source or on more powerful, distant nodes~\cite{Ramezani23}.

However, scheduling these workflows in such a dynamic and heterogeneous environment is a significant challenge. Traditional approaches often rely on heuristic algorithms like Heterogeneous Earliest Finish Time (HEFT)~\cite{Topcuoglu2002}. While effective in static environments, the fixed rules of these heuristics limit their adaptability to runtime dynamics, such as fluctuating network conditions. Consequently, Deep Reinforcement Learning (DRL) has emerged as a powerful paradigm for developing adaptive scheduling policies. Various DRL methods, including Deep Q-Networks (DQN)~\cite{Tang22,Swarup21} and policy-gradient algorithms like Deep Deterministic Policy Gradient (DDPG)~\cite{Cheng24,Fan20}, have been explored, along with advanced structures like Hierarchical Reinforcement Learning (HRL)~\cite{Jayanetti22,Lei22} that better align with the layered infrastructure.

Despite its promise, efficiently scheduling sequential medical workflows across this heterogeneous infrastructure remains a critical challenge, stemming from two fundamental issues. First, the scheduling problem is \textbf{NP-hard}, requiring the scheduler to navigate the inherent \textbf{computation-communication trade-off} at each step. As formally defined in our problem formulation (Section 2.2), assigning a task to a powerful but remote node reduces computation time but incurs significant communication latency, making optimal placement computationally intractable. Second, while DRL is a promising paradigm, conventional \textbf{``flat'' DRL schedulers} are ill-suited for this hierarchical environment. Treating all nodes across the cloud, fog, and edge layers as a single, monolithic action space fails to exploit the system's natural structure. This leads to an exponentially large action space, resulting in \textbf{inefficient exploration, poor scalability}, and \textbf{slow convergence}, which hinders the discovery of an effective scheduling policy.

To address these challenges, we propose a \textbf{Two-tier DDPG-based scheduling framework}. This framework responds to these issues by decomposing the complex scheduling decision into a hierarchical process. A high-level \textbf{global controller} performs strategic \textbf{layer selection} (edge, fog, or cloud), explicitly managing the computation-communication trade-off. Concurrently, specialized \textbf{local controllers} handle fine-grained \textbf{node assignment} within the chosen layer, effectively taming the large action space and ensuring scalability. By aligning the learning architecture with the physical infrastructure, our approach enables efficient and robust policy learning for makespan minimization.

The main contributions of this work are summarized as follows:
\begin{enumerate}
\item We propose a cloud-fog-edge collaborative computing framework for MIoT workflows that performs sequential medical tasks with heterogeneous computational and communication requirements. We formally formulate the workflow scheduling as a constrained optimization problem that minimizes makespan while satisfying resource capacity, task precedence, and memory feasibility constraints across the three-layer infrastructure.
\item We propose a two-tier DDPG scheduler that solves the formulated NP-hard scheduling problem through decomposed decision-making: a global controller for layer selection and specialized local controllers for node assignment within each layer. This two-tier architecture naturally aligns with the physical infrastructure while enabling efficient policy learning.
\item We conduct extensive empirical validation, demonstrating that our adaptive scheduler achieves increasingly superior performance over myopic online baselines as workflow complexity rises, highlighting its ability to learn effective long-term strategies.
\end{enumerate}

The remainder of this paper is structured as follows: Section~\ref{sec:system_design} presents the system design and problem formulation. Section~\ref{sec:methodology} details the two-tier DDPG scheduling methodology. Section~\ref{sec:experiments} provides the experimental evaluation and results. Finally, Section~\ref{sec:conclusion} concludes the paper.

%\vspace{-0.4cm}
\section{System Design}
\label{sec:system_design}
This section formally presents the details of Two-tier cloud-fog-edge collaborative computing
framework % 添加本章纲要
and defines the workflow scheduling problem as a constrained optimization problem, analyzes its inherent complexity, and establishes the motivation for employing a learning-based approach.
%\vspace{-0.4cm}
%This section first presents our proposed Two-tier Three-layer Computing Framework (TTCF). It then provides a detailed mathematical model of the system, culminating in a formal definition of the constrained optimization problem for workflow scheduling.
%\vspace{-0.4cm}

%删除overview,将overview内容直接置于第二章节下，总结本章节内容，引出下文内容
%\vspace{-0.4cm}
% 顶层包装与核心叙事 (Overall Packaging & Narrative)研究问题: 在异构的云-雾-边（Cloud-Fog-Edge）基础设施上，为序列化的医疗物联网（MIoT）工作流进行任务调度，核心目标是最小化端到端完成时间（Makespan）。问题定性: 这是一个  的组合优化问题。核心挑战与动机:静态启发式算法的局限性: HEFT等传统算法在动态、不确定的MIoT环境中适应性差。two-tier motivation: 直接将所有节点视为一个大的动作空间，会导致探索效率低下、收敛缓慢且扩展性差，分治的策略降低了策略复杂度。
\subsection{System Framework}

\begin{figure}[htbp]
\centering
\includegraphics[width=0.85\linewidth]{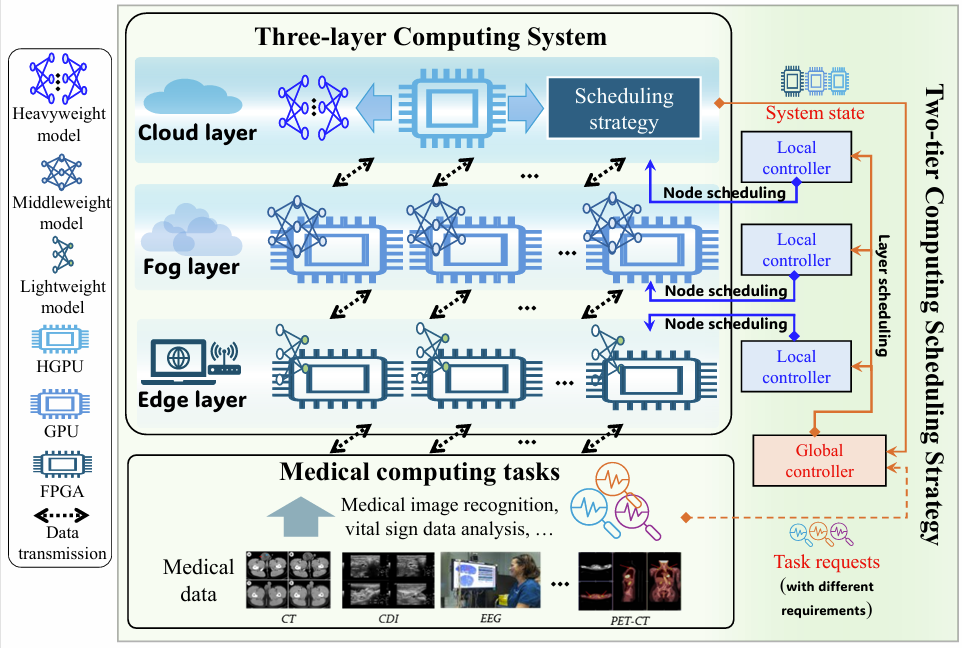} 
\caption{The two-tier DDPG-based scheduling framework for processing sequential MIoT task workflows in cloud-fog-edge infrastructures.}
\label{fig:system_architecture}
\end{figure}

As depicted in Fig.~\ref{fig:system_architecture}, our proposed system is a Two-tier cloud-fog-edge collaborative computing framework for efficiently processing medical tasks. It comprises three main components: i) diverse medical task workflows, ii) a cloud-fog-edge collaborative computing infrastructure, and iii) a two-tier (global and local) scheduling strategy.

\subsubsection{Medical Computing Task Workflows}
MIoT workflows consist of tasks with extremely heterogeneous resource demands, from lightweight signal processing to computationally intensive analytics~\cite{Alatoun22}. We model such a workflow as an ordered sequence $\mathcal{V} = (v_1, \dots, v_T)$, where each task $v$ is defined by a tuple $(w_v, m_v, d_v)$ representing its computational workload (\textsf{MI}), memory requirement (\textsf{MB}), and output data size (\textsf{MB}). Scheduling these workflows to minimize the end-to-end makespan demands constant navigation of the trade-off between the high computational power of upper tiers and their associated communication latency. As static heuristics are ill-equipped for this dynamic challenge, we propose a Two-tier DDPG framework. This approach leverages deep reinforcement learning to learn an adaptive, hierarchical policy that dynamically manages the computation-communication trade-off, overcoming the limitations of non-adaptive schedulers.

\subsubsection{Cloud-Fog-Edge Collaborative Computing Framework}
We consider a cloud-fog-edge framework organized as a three-layer hierarchy of heterogeneous resources, indexed by \(\ell \in \{l_{\text{edge}}, l_{\text{fog}}, l_{\text{cloud}}\}\). Each layer $\ell$ contains a set of nodes $\mathcal{N}_\ell$. The edge layer ($l_{\text{edge}}$), closest to data sources, features resource-constrained nodes (e.g., FPGAs) for low-latency tasks. The intermediate fog layer ($l_{\text{fog}}$) provides more powerful nodes (e.g., GPUs) for moderate workloads, while the centralized cloud layer ($l_{\text{cloud}}$) offers high-performance nodes (e.g., HGPUs) for the most complex tasks. This structure creates a fundamental trade-off: accessing the superior computational power of upper layers incurs greater communication latency. The set of all nodes is $\mathcal{N} = \mathcal{N}_{l_{\text{edge}}} \cup \mathcal{N}_{l_{\text{fog}}} \cup \mathcal{N}_{l_{\text{cloud}}}$, where each node \(n \in \mathcal{N}_\ell\) is characterized by its computational capacity $C_n$ (\textsf{MIPS}) and memory capacity $M_n$ (\textsf{MB}).

\subsubsection{The Two-tier DDPG Scheduling Framework}
\label{subsubsec: two-tier}
Scheduling diverse MIoT workflows across the cloud-fog-edge infrastructure is challenging due to: i) the need to manage complex precedence constraints and heterogeneous workloads, demanding a long-term resource allocation perspective; and ii) the difficulty in tracking the dynamic state of the entire network, which hinders timely decision-making. To address this, we employ the Two-tier DDPG Scheduling Framework, which offers flexibility in response to dynamic system states and task requirements. This architecture comprises a \textbf{global controller} (tier-1) for high-level layer selection and multiple \textbf{local controllers} (tier-2) for fine-grained node assignment. The global controller allocates each task to the most suitable layer, after which the corresponding local controller selects a specific node within that layer. The following sections will formalize this process as a constrained optimization problem.

%核心任务是将前面章节描述的系统和挑战，转为数学模型，确立了研究的最终目标 (Objective)：最小化工作流的总执行时间，即 Makespan，次强调了问题的内在复杂性，决策的依赖性：当前对一个任务的调度决策，会直接影响并限制后续所有任务的调度选择，核心的权衡：再次重申了在云-雾-边架构中，选择计算能力更强的节点必然会带来更高的通信延迟，这个矛盾是无法回避的。

\subsection{Problem Formulation of Scheduling Strategy}

Our primary objective is to minimize the total execution time, or \textbf{makespan}, for a sequential MIoT workflow. This is complex due to interdependent assignment decisions, where each choice impacts all subsequent options and must navigate the fundamental computation-communication trade-off. Assigning a task to a powerful but remote node may reduce computation time, but the incurred data transfer latency can nullify this advantage. To formalize this, we first model task execution and data transmission costs, then construct the makespan objective and define the constrained optimization problem.

%定义基础模型，通信时间模型（公式 \ref{eq:comm_time}）：将数据从一层传输到另一层的时间，由固定的传输延迟和与数据大小相关的传输时间构成，计算时间模型：一个任务在某个节点上的执行时间，由任务的总计算量除以该节点的计算能力决定。分场景建模：基于上述基础模型，本节详细分析了将一个任务分配到边、雾、云三个不同层次时，其总时间成本的构成，Edge: 只有计算时间。Fog: 包含一次通信时间（Edge到Fog）和计算时间。Cloud: 包含两次通信时间（Edge到Fog，再到Cloud）和计算时间。统一成本函数：将上述三种情况统一成一个单一的、分段的成本函数 T_cost

\subsubsection{Preliminary modeling} We use $l_{\text{up}}$ to denote the upstream layer and $l_{\text{down}}$ to denote the downstream layer. Herein, the inter-layer latency and bandwidth are given by $T_{\mathrm{tr}}(l_{\text{up}}, l_{\text{down}})$ and $B(l_{\text{up}}, l_{\text{down}})$, respectively. In particular, communication is always initiated from the lower layer to the upper layer and is independent of nodes. Typically, MIoT performs diverse computing tasks (e.g., medical image recognition, vital sign data analysis), where the input of one task is the result of another task. 
Let $\mathcal{V} = (v_1, v_2, \dots, v_T)$ denote an ordered sequence of tasks, where $T$ is the number of tasks.

%Each task $v$ is characterized by a three-item tuple $(w_v, m_v, d_v)$, where $w_v$ is its computational workload, $m_v$ is its memory requirement, and $d_v$ is its output data size. 

To better describe our proposed system, here, we define the total communication between $l_{\text{down}}$ and $l_{\text{up}}$ with transmitting task $v$ as:
\begin{equation}
\tau_{\text{comm}}(l_{\text{down}}, l_{\text{up}}, v) =T_{\mathrm{tr}}(l_{\text{up}}, l_{\text{down}}) + \frac{d_v}{B(l_{\text{down}}, l_{\text{up}})}.
\label{eq:comm_time}
\end{equation}

The time to process task $v$ on a given node $n$ is its execution time, denoted as $t_{\text{exec}}(v, n) = w_v/C_n$. At the beginning, the global controller assigns each task $v$ to a specific node $n$, leading to the following three cases:
\begin{enumerate}
    \item \textbf{Edge execution.}  
    If a node in the edge layer$\mathcal{N}_{l_{\text{edge}}}$ is selected, the task is directly processed on this node. The total latency is purely the execution time
    \[
    t_{\text{edge}}(v) = t_{\text{exec}}(v, n_{{\text{edge}}}),
    \]
    since no inter-layer communication is required. Upon completion, the output is immediately forwarded to the subsequent task in the workflow.
    
    \item \textbf{Fog execution.}  
    If a node in the fog layer ($\mathcal{N}_{l_{\text{fog}}}$) is selected, the task $v$ first arrives at an edge node and is then transmitted to a fog node. The total delay includes both the communication cost and execution cost:
    \[
    t_{\text{fog}}(v) = \tau_{\text{comm}}(l_{\text{edge}}, l_{\text{fog}}, v) + t_{\text{exec}}(v, n_{{\text{fog}}}).
    \]
    This reflects the extra time required for transmitting intermediate results from the edge layer to the fog layer before computation can begin.
    
    \item \textbf{Cloud execution.}  
    If the controller assigns the task to a node in the cloud layer ($\mathcal{N}_{l_{\text{cloud}}}$), the task output must be sequentially transmitted from the edge layer through the fog layer to the cloud layer. Thus, the communication delay is cumulative across two layer, followed by the execution at the cloud:
    \[
    t_{\text{cloud}}(v) = \tau_{\text{comm}}(l_{\text{edge}}, l_{\text{fog}}, v) + \tau_{\text{comm}}(l_{\text{fog}}, l_{\text{cloud}}, v) + t_{\text{exec}}(v, n_{{\text{cloud}}}).
    \]
    Although this path incurs the largest communication overhead, the cloud’s powerful computational capacity often yields the lowest execution time per task.
\end{enumerate}

These three distinct scheduling scenarios can be formally consolidated into a unified mathematical model. For convenience in the subsequent formulation, we define the total time cost for assigning a task $v$ to a specific computing node $n$, denoted as $T_{\text{cost}}(v, n)$. The cost depends on the layer affiliation of the node, i.e., $n \in \mathcal{N}_{l_{\text{edge}}}, \mathcal{N}_{l_{\text{fog}}}, \mathcal{N}_{l_{\text{cloud}}}\}$. Accordingly, the total cost $T_{\text{cost}}(v, n)$ is formulated as the following piecewise function: $T_{\text{cost}}(v, n) =$
\begin{equation}
\begin{cases}
    t_{\text{exec}}(v, n), 
    & \text{if } n \in \mathcal{N}_{l_{\text{edge}}}, \\[8pt]
    \tau_{\text{comm}}(l_{\text{edge}}, l_{\text{fog}}, v) + t_{\text{exec}}(v, n), 
    & \text{if } n \in \mathcal{N}_{l_{\text{fog}}}, \\[8pt]
    \tau_{\text{comm}}(l_{\text{edge}}, l_{\text{fog}}, v) 
    + \tau_{\text{comm}}(l_{\text{fog}}, l_{\text{cloud}}, v) 
    + t_{\text{exec}}(v, n), 
    & \text{if } n \in \mathcal{N}_{l_{\text{cloud}}}.
\end{cases}
\label{eq:task_cost}
\end{equation}
This formulation encapsulates the fundamental trade-off between computation and communication that the scheduler must navigate: while upper-layer nodes provide stronger computational capabilities and lower execution time, they inevitably incur additional communication overhead.

%定义决策变量：首先，引入了核心的二元决策变量 x_vn。这个变量是连接抽象问题与具体数学解法的桥梁，它的取值（0或1）直接代表了“是否将任务v分配给节点n”这一决策，分配唯一性约束，内存可行性约束，决策变量约束

\subsubsection{Optimization problem.}
For each task $v\in\mathcal{V}$ and node \(n \in \mathcal{N}_\ell\), we introduce a binary decision variable $x_{v,n}\in\{0,1\}$, where $x_{v,n}=1$ if task $v$ is assigned to node $n$, and 0 otherwise. Since tasks execute sequentially (per the workflow order), the workflow latency equals the sum of the per-task costs. Our objective is to minimize the end-to-end workflow latency (equivalently, the makespan for a strictly sequential pipeline) subject to assignment and feasibility constraints:
\begin{align}
\min_{\{x_{v,n}\}} \quad & T_{\text{makespan}} = \sum_{v\in\mathcal{V}} \sum_{n\in\mathcal{N}} x_{v,n}\, T_{\mathrm{cost}}(v,n) \label{eq:obj}\\
\text{s.t.}\quad 
& \sum_{n\in\mathcal{N}} x_{v,n}=1, \quad \forall v\in\mathcal{V}, \label{eq:assign_once}\\
& m_v\, x_{v,n} \le M_n, \quad \forall v\in\mathcal{V},~\forall n\in\mathcal{N}, \label{eq:mem_feas}\\
& x_{v,n}\in\{0,1\}, \quad \forall v\in\mathcal{V},~\forall n\in\mathcal{N}. \label{eq:binary}
\end{align}

\noindent
Constraint~\cref{eq:assign_once} enforces that each task is executed exactly once; 
\cref{eq:mem_feas} ensures memory feasibility on the selected node;
and~\cref{eq:binary} specifies integrality.
Because $\mathcal{V}$ is a strictly ordered sequence, precedence is implicit and the end-to-end latency equals the sum in~\cref{eq:obj}.
Problem~\cref{eq:obj}--\cref{eq:binary} captures the core computation--communication trade-off: upper layers offer larger computational capacity $C_n$ but incur additional inter-layer communication as encoded in $T_{\mathrm{cost}}(v,n)$. This problem is \textbf{NP-hard}, as it can be reduced to the Generalized Assignment Problem (GAP), a well-known combinatorial optimization challenge~\cite{Garey1979}. Specifically, assigning tasks (items) from $\mathcal{V}$ to heterogeneous nodes (bins) in $\mathcal{N}$ while respecting memory constraints \cref{eq:mem_feas} and minimizing the total cost \cref{eq:obj} makes it computationally intractable to find an optimal solution for large-scale instances, thus motivating our learning-based approach.

% ===================== HD-DDPG-based Two-tier Scheduling =====================
\section{Two-tier DDPG-based Scheduling}
\label{sec:methodology}
%阐述了我们如何利用强化学习来解决第二章中定义的动态的调度问题。为此，我们首先将问题建模为一个马尔可夫决策过程（MDP），其核心贡献在于将“两层式”思想完全融入其中：设计了分层的状态空间（宏观层级统计/微观节点信息）和动作空间（全局选层/局部选点），并构建了与Makespan目标直接挂钩的奖励函数。在此基础上，我们提出了一个同样分层的两层式DDPG算法架构，由一个全局控制器进行战略性的层级权衡，多个局部控制器进行战术性的节点负载均衡。最后，本章通过策略优化的数学公式和一份完整的伪代码算法，展示了这两个层级的控制器如何通过共享的奖励信号进行联合训练，最终学习到一个能够最小化端到端延迟的、鲁棒的调度策略。
\subsection{Reinforcement Learning Formulation}

The dynamic nature of MIoT environments, characterized by runtime uncertainties like fluctuating bandwidth, renders static heuristics ineffective. To overcome this, we formulate the scheduling task as a sequential decision-making problem and employ the \textbf{Two-tier DDPG framework} from Section~\ref{subsubsec: two-tier}. This framework decouples the assignment decision into two hierarchical steps: a tier-1 \textbf{global controller} performs layer selection to manage the core computation-communication trade-off, and a tier-2 \textbf{local controller} handles node assignment for load balancing within the chosen layer. This hierarchical decomposition mirrors the physical infrastructure, simplifying the learning problem and enabling the discovery of a robust policy for minimizing the makespan.

\subsection{MDP Modeling}
%移除了“利用率”、“负载均衡”等前文未铺垫的新概念，用“计算负载指标”、“选择最早完成任务的节点”等更精确且直接与优化目标（最小化完成时间）相关的描述进行替换；修正了将“总内存”与“可用内存”混淆的逻辑瑕疵；并显式地将状态定义与第二章提出的“网络动态性”挑战进行挂钩

%把我们的调度问题严格地映射为一个马尔可夫决策过程 (MDP)。MDP是强化学习的通用数学语言，我们必须定义好它的四个核心要素：状态 (S)、动作 (A)、转移 (P) 和奖励 (R)。

We model the sequential task assignment problem as a Markov Decision Process (MDP) defined by the tuple $\langle \mathcal{S}, \mathcal{A}, \mathcal{P}, \mathcal{R} \rangle$. At each decision step $t$ for a task $v_t$, the scheduler observes the current system state $s_t \in \mathcal{S}$, selects an action $a_t \in \mathcal{A}$, receives a reward $r_t$ from the reward function $\mathcal{R}$, and the system transitions to the next state $s_{t+1}$ according to the dynamics $\mathcal{P}$. The framework's hierarchical design is explicitly embedded within each component of this MDP.

\subsubsection{State Space \(\mathcal{S}\)}
The state representation at decision step \(t\) follows a two-tier hierarchy. The \textbf{global state} \(S_{\text{global}}\) captures aggregated layer-level statistics to inform the high-level layer selection, including the average computational load across each layer, the aggregated available memory capacity for each layer ($\sum_{n \in \mathcal{N}_\ell} M_n$), inter-layer communication queue lengths, and the current task's characteristics $(w_{v_t}, m_{v_t}, d_{v_t})$. For each layer \(\ell\), the \textbf{local state} \(S_{\text{local}}^{(\ell)}\) provides fine-grained node-level features for optimal node selection, encompassing individual node residual memory, the expected available time for each node based on its current task queue, and node-specific computational capacity $C_n$.

\subsubsection{Action Space \(\mathcal{A}\)}
Each scheduling action is hierarchically decomposed as \(a_t = (\ell, n)\), where the scheduler first selects a target layer \(\ell\) and then a specific node \(n \in \mathcal{N}_\ell\). This two-stage action directly maps to the binary decision variables in our optimization problem. Executing action \((\ell, n)\) for task $v_t$ is equivalent to setting $x_{v_t,n} = 1$ and $x_{v_t,m} = 0$ for all other nodes $m \neq n$. This mapping ensures that assignments satisfy constraint \cref{eq:assign_once}, guaranteeing each task is executed exactly once. Memory feasibility (constraint \cref{eq:mem_feas}) is enforced by masking out nodes with insufficient memory from the action space.

\subsubsection{State Transitions \(\mathcal{P}\)}
Given the current state $s_t$ and action $a_t = (\ell, n)$, the next state $s_{t+1}$ is deterministically computed using our latency models from \cref{eq:comm_time}--\cref{eq:task_cost}. The transition process updates the assigned node's temporal availability by adding the total cost $T_{\text{cost}}(v_t, n)$ to its expected completion time and reduces its available memory by $m_{v_t}$. These changes are then propagated to update layer-level aggregated statistics. Finally, the system advances to the next task $v_{t+1}$, and since tasks follow a strict sequential order, precedence constraints are implicitly satisfied.

\subsubsection{Reward Function \(\mathcal{R}\)}
The reward function is designed to align with our optimization objective, incorporating auxiliary terms to facilitate learning. The core component directly addresses makespan minimization:
\begin{equation}
r_t = - \sum_{n\in\mathcal{N}} x_{v_t,n} \cdot T_{\mathrm{cost}}(v_t,n) + \beta_1 \cdot r_{\text{bonus}} + \beta_2 \cdot r_{\text{eff}}.
\label{eq:rl_reward_detailed}
\end{equation}
The first term directly penalizes the delay incurred by assigning task $v_t$, establishing a correspondence with the objective in \cref{eq:obj}. Since only one $x_{v_t,n}$ is 1, this penalty equals the actual cost $T_{\mathrm{cost}}(v_t,n)$ of the assignment. The auxiliary terms serve complementary roles: $r_{\text{bonus}}$ provides a positive reward upon successful workflow completion, and $r_{\text{eff}}$ rewards efficient resource utilization. The coefficients $\beta_1$ and $\beta_2$ are tuned to maintain focus on the primary objective while providing sufficient learning signals. This structure ensures that maximizing the cumulative reward implicitly minimizes the workflow makespan.
\subsection{Two-tier DDPG Architecture}

Our Two-tier DDPG framework employs two coordinated DDPG controllers that operate hierarchically:

\subsubsection{Global Controller}
The global controller consists of an actor--critic pair $(\mu_{\theta^{G}}, Q_{\phi^{G}})$ that processes the global state $S_{\text{global}}$ to select the optimal layer $\ell$. The actor network $\mu_{\theta^{G}}: S_{\text{global}} \rightarrow \mathbb{R}^3$ outputs a probability distribution over the cloud-fog-edge, while the critic network $Q_{\phi^{G}}: S_{\text{global}} \times \{l_{\text{edge}}, l_{\text{fog}}, l_{\text{cloud}}\} \rightarrow \mathbb{R}$ evaluates the Q-value of layer selection decisions. This controller learns to balance the fundamental trade-off between computation speed (favoring upper layers) and communication overhead (favoring lower layers).

\subsubsection{Local Controller}
For each layer $\ell$, a dedicated local controller with actor--critic pair $(\mu_{\theta^{\ell}}, Q_{\phi^{\ell}})$ maps the local state $S_{\text{local}}^{(\ell)}$ to a specific node $n \in \mathcal{N}_\ell$. The actor network $\mu_{\theta^{\ell}}: S_{\text{local}}^{(\ell)} \rightarrow \mathbb{R}^{|\mathcal{N}_\ell|}$ generates a continuous action vector over feasible nodes, while the critic $Q_{\phi^{\ell}}$ evaluates node-level decisions. Each local controller specializes in load balancing and resource optimization within its respective layer.The two controllers act sequentially—global followed by local—but are trained jointly using the shared reward signal \cref{eq:rl_reward_detailed}. This joint training ensures that both levels of the hierarchy learn complementary policies that collectively optimize the end-to-end makespan.

\subsection{Policy Optimization}
\label{subsec: policy}
%核对了代码与此章节的定义，公式，表述，确保逻辑一致。
\noindent\textbf{Mini-batch Sampling.}
At each gradient step, we uniformly sample a mini-batch of size \(B\) from the replay buffer $\mathcal{D}$, i.e., $\big\{(s_t^i, a_t^i, r_t^i, s_{t+1}^i)\big\}_{i=1}^B \sim \mathcal{D}$, where the action is a tuple $a_t^i=(\ell_t^i, n_t^i)$. Feasible-node masking, which enforces memory constraint \cref{eq:mem_feas}, is applied by filtering the action space to the set $\mathcal{F}_t^i = \{\, n \in \mathcal{N}_{\ell_t^i}\ :\ m_{v_t^i} \le M_n \,\big\}$.

\noindent\textbf{Exploration.} During data collection, hierarchical actions are perturbed by zero-mean Gaussian noise ($\epsilon_t^G, \epsilon_t^{\ell_t}$) with annealing to gradually reduce exploration, as shown in \cref{eq:global_noise_addon} and \cref{eq:local_noise_addon}.
% --- Start of side-by-side equations for Exploration ---
\vspace{-0.3cm}
\begin{figure}[htbp]
\centering
\begin{minipage}{0.48\linewidth}
\small
\begin{equation}
\ell_t = \arg\max_{\ell} 
\mu_{\theta^{G}}(S_{\text{global},t}) + \epsilon_t^{G}
\label{eq:global_noise_addon}
\end{equation}
\end{minipage}
\hfill
\begin{minipage}{0.48\linewidth}
\small
\begin{equation}
n_t = \arg\max_{n \in \mathcal{F}_t} 
\mu_{\theta^{\ell_t}}(S_{\text{local},t}^{(\ell_t)}) + \epsilon_t^{\ell_t}
\label{eq:local_noise_addon}
\end{equation}
\end{minipage}
\end{figure}
% --- End of side-by-side equations ---

\vspace{-0.3cm}

\noindent\textbf{Target Actions \& TD Targets.} For each batch element \(i\), the target actors propose next-step greedy actions under the masked feasible set:
\begin{align}
\tilde{\ell}_{t+1}^i \;&=\; \arg\max_{\ell} \mu_{\theta^{G'}}\!\big(S_{\text{global},t+1}^i\big), \label{eq:target_layer_addon}\\
\tilde{n}_{t+1}^i \;&=\; \arg\max_{n \in \mathcal{N}_{\tilde{\ell}_{t+1}^i}\cap \mathcal{F}_{t+1}^i} 
\mu_{\theta^{\tilde{\ell}_{t+1}^i{}'}}\!\big(S_{\text{local},t+1}^{(\tilde{\ell}_{t+1}^i),i}\big). \label{eq:target_node_addon}
\end{align}
\vspace{-0.5cm}
The 1-step TD targets for global and local critics are then calculated as:
% --- Start of side-by-side equations for TD Targets ---
\begin{figure}[htbp]
\centering
\begin{minipage}{0.48\linewidth}
\small
\begin{equation}
y_G^i = r_t^i + \gamma\, Q_{\phi^{G'}}\!\big(s_{t+1}^i,\, \tilde{\ell}_{t+1}^i\big)
\label{eq:td_target_global_addon}
\end{equation}
\end{minipage}
\hfill
\begin{minipage}{0.48\linewidth}
\small
\begin{equation}
y_{\ell}^i = r_t^i + \gamma\, Q_{\phi^{\ell'}}\!\big(s_{t+1}^i,\, \tilde{n}_{t+1}^i\big)
\label{eq:td_target_local_addon}
\end{equation}
\end{minipage}
\end{figure}%
\vspace{-0.5cm}
% --- End of side-by-side equations ---
where the local target $y_{\ell}^i$ is computed for the layer $\ell=\ell_t^i$ chosen at step $t$.

\noindent\textbf{Critic Updates (Mini-batch MSE).} Global and local critics are updated by minimizing their respective mean squared TD errors. The total critic loss is the sum of these components, 
\small
$\mathcal{L}_{\text{critic}}=\mathcal{L}_{\text{critic}}^{G}+\sum_{\ell}\mathcal{L}_{\text{critic}}^{\ell}$, with each component defined as:
\begin{align}
\small
\mathcal{L}_{\text{critic}}^{G}(\phi^{G}) 
&= \frac{1}{B}\sum_{i=1}^{B}\Big(y_{G}^i - Q_{\phi^{G}}(s_t^i,\ell_t^i)\Big)^2,
\label{eq:critic_loss_global_addon}\\
\mathcal{L}_{\text{critic}}^{\ell}(\phi^{\ell})
&= \frac{1}{B}\sum_{i=1}^{B} \mathbf{1}\{\ell_t^i=\ell\} \Big(y_{\ell}^i - Q_{\phi^{\ell}}(s_t^i,n_t^i)\Big)^2.
\label{eq:critic_loss_local_addon}
\end{align}

\noindent\textbf{Actor Updates.} Actors ascend their critics’ Q-values. The policy gradient is taken with respect to the continuous actor output, which we define as $a \triangleq \mu_{\theta}(s)$, with batch instances $a_t^{i,G} = \mu_{\theta^{G}}(S_{\text{global},t}^i)$ and $a_t^{i,\ell} = \mu_{\theta^{\ell}}(S_{\text{local},t}^{(\ell),i})$. The gradients for the global and local actors are given by:
\begin{align}
\small
\nabla_{\theta^{G}} J^{G}(\theta^{G})
&=\frac{1}{B}\sum_{i=1}^{B}
\Big[
\nabla_{a}\, Q_{\phi^{G}}\!\big(s_t^i,a\big)\big|_{a=a_t^{i,G}}
\cdot
\nabla_{\theta^{G}} \mu_{\theta^{G}}(S_{\text{global},t}^i)
\Big],
\label{eq:actor_update_global_addon} \\
\nabla_{\theta^{\ell}} J^{\ell}(\theta^{\ell})
&=\frac{1}{B}\sum_{i=1}^{B}\mathbf{1}\{\ell_t^i=\ell\}
\Big[
\nabla_{a}\, Q_{\phi^{\ell}}\!\big(s_t^i,a\big)\big|_{a=a_t^{i,\ell}}
\cdot
\nabla_{\theta^{\ell}} \mu_{\theta^{\ell}}\!\big(S_{\text{local},t}^{(\ell),i}\big)
\Big].
\label{eq:actor_update_local_addon}
\end{align}

\noindent\textbf{Target Network Soft Updates.} After each gradient step, target networks are updated via Polyak averaging with $\tau\in(0,1]$. This is applied to all global and local parameters according to the rule in \cref{eq:soft_update_addon}.
\begin{equation}
\theta' \leftarrow \tau \theta + (1-\tau)\theta', \quad \phi' \leftarrow \tau \phi + (1-\tau)\phi'
\label{eq:soft_update_addon}
\end{equation}

\subsection{Training and Scheduling Algorithm}

\cref{alg:hdddpg} presents the Two-tier DDPG training procedure. The algorithm employs hierarchical decision-making where a global controller selects the execution layer and local controllers assign specific nodes. Both tiers are trained jointly through shared reward signals to minimize workflow makespan.

\begin{algorithm}[t]
\caption{Two-tier DDPG Scheduling}
\label{alg:hdddpg}
\begin{algorithmic}[1]
\Require Infrastructure $\{\mathcal{N}_{\ell}\}_{\ell\in\{l_{\text{edge}},l_{\text{fog}},l_{\text{cloud}}\}}$, hyperparameters $(\gamma,\tau,B)$
\Ensure Trained policy $\pi=(\mu_{\theta^{G}},\{\mu_{\theta^{\ell}}\})$
\State Initialize global actor--critic $(\mu_{\theta^{G}},Q_{\phi^{G}})$ and targets
\State Initialize local actor--critics $\{(\mu_{\theta^{\ell}},Q_{\phi^{\ell}})\}_{\ell}$ and targets
\State Initialize replay buffer $\mathcal{D}$
\For{episode $=1$ to $N_{\text{episodes}}$}
  \State Sample workflow $\mathcal{V}=(v_1,\dots,v_T)$; reset env. state
  \For{task $v_t \in \mathcal{V}$ (sequential order)}
    \State \textbf{Global decision:}
    \State \quad $a_t^G \leftarrow \mu_{\theta^{G}}(S_{\text{global},t})$ \Comment{Get continuous action vector}
    \State \quad $\ell_t \leftarrow \arg\max_{\ell} (a_t^G + \epsilon_t^G)$ \Comment{Select discrete action via noisy argmax}
    \State \textbf{Local decision:}
    \State \quad $a_t^{\ell_t} \leftarrow \mu_{\theta^{\ell_t}}(S_{\text{local},t}^{(\ell_t)})$
    \State \quad $\mathcal{F}_t \leftarrow \{n\in\mathcal{N}_{\ell_t} \mid m_{v_t}\le M_n\}$ \Comment{Mask from \cref{eq:mem_feas}}
    \State \quad $n_t \leftarrow \arg\max_{n\in\mathcal{F}_t} (a_t^{\ell_t} + \epsilon_t^{\ell_t})$
    \State \textbf{Execute:} assign $x_{v_t,n_t}\!=\!1$; obtain cost $T_{\text{cost}}(v_t,n_t)$ via \cref{eq:task_cost} 
    \State \textbf{Reward:} compute $r_t$ using \cref{eq:rl_reward_detailed}
    \State Observe $s_{t+1}$; store $(s_t, (a_t^G, a_t^{\ell_t}), r_t, s_{t+1})$ in $\mathcal{D}$ \Comment{Store continuous actions}
    \If{$|\mathcal{D}|\ge B$ \textbf{and} update step}
      \State Sample mini-batch from $\mathcal{D}$
      \State Build TD targets using \cref{eq:target_layer_addon}--\cref{eq:td_target_local_addon}
      \State Update critics by minimizing \cref{eq:critic_loss_global_addon} and \cref{eq:critic_loss_local_addon}
      \State Update actors using \cref{eq:actor_update_global_addon} and \cref{eq:actor_update_local_addon}
      \State Soft-update targets by \cref{eq:soft_update_addon}
    \EndIf
  \EndFor
\EndFor
\State \Return $\pi=(\mu_{\theta^{G}},\{\mu_{\theta^{\ell}}\})$
\end{algorithmic}
\end{algorithm}
The algorithm begins by initializing the hierarchical network architecture (Lines 1-3). During each episode, tasks are processed sequentially. For each task $v_t$, the global controller selects a layer $\ell_t$ via a noisy argmax operation on its continuous output (Lines 8-9). Subsequently, the corresponding local controller selects a node $n_t$ from the memory-feasible set $\mathcal{F}_t$ using the same mechanism (Lines 11-13). The transition, including the continuous action vectors required for gradient calculation, is then stored in the replay buffer $\mathcal{D}$ (Line 16).
Policy optimization occurs periodically by sampling a mini-batch from $\mathcal{D}$ (Lines 17-23). Both critic and actor networks are updated using the temporal difference errors and policy gradients formulated in \cref{subsec: policy}. Target networks are softly updated with parameter $\tau$ to maintain training stability. Through this iterative process, the agent learns a hierarchical policy that effectively balances computation-communication trade-offs while respecting system constraints.
\section{Experiments and Results}
\label{sec:experiments}

This section presents a comprehensive evaluation of the proposed Two-tier DDPG framework. The experiments are designed to validate the framework's effectiveness in solving the MIoT scheduling problem formulated in \cref{eq:obj}--\cref{eq:binary}. We first analyze the training process, demonstrating that maximizing the cumulative reward from \cref{eq:rl_reward_detailed} leads to a stable policy that minimizes the makespan. Subsequently, the converged policy's makespan performance is benchmarked against established heuristics across workflows of varying complexity. For full transparency, our source code and experiment data are publicly available on GitHub~\cite{Fu24_github}.

\subsection{Experimental Design}
\label{sec:exp_design}

We evaluate the proposed Two-tier DDPG scheduler on sequential MIoT workflows ($\mathcal{V}$) of varying complexity, using the workflow \textbf{makespan} ($T_{\text{makespan}}$) from \cref{eq:obj} as the primary performance metric. Its performance is benchmarked against four algorithms: the powerful \textbf{HEFT} heuristic~\cite{Topcuoglu2002}, a myopic online \textbf{Greedy} scheduler minimizing immediate cost (\cref{eq:task_cost}), a round-robin \textbf{FCFS} policy, and a \textbf{Random} baseline. Workflows are synthetically generated across four difficulty levels (L1--L4), defined by an increasing number of tasks ($|\mathcal{V}|$): \textbf{L1} (5--8), \textbf{L2} (9--12), \textbf{L3} (13--18), and \textbf{L4} (19--25).Experiments were conducted on a Python-based discrete-event simulator implementing the framework from Section~\ref{sec:system_design}. The simulated infrastructure (Table~\ref{tab:infra_params}) features three heterogeneous layers with distinct communication links: Edge-to-Fog (10~ms, 200~\textsf{Mbps}) and Fog-to-Cloud (40~ms, 100~\textsf{Mbps}).

\begin{table}[htbp]
\centering
\caption{Simulated Cloud-Fog-Edge Infrastructure Parameters}
\label{tab:infra_params}
\begin{tabular}{@{}l l c c c@{}}
\toprule
\textbf{Layer} & \textbf{Node} & \textbf{Count} & \textbf{Capacity ($C_n$, MIPS)} & \textbf{Memory ($M_n$, MB)} \\ \midrule
Edge   & FPGA & 4 & 800 -- 1,200   & 2,048 \\
Fog    & GPU  & 3 & 2,500 -- 3,000 & 6,144 -- 8,192 \\
Cloud  & HGPU & 1 & 8,000          & 32,768 \\ \bottomrule
\end{tabular}
\end{table}

The Two-tier DDPG scheduler (Section~\ref{sec:methodology}) was trained for 700 episodes. Key hyperparameters for policy optimization are listed in Table~\ref{tab:hdddpg_hyperparams}.

\begin{table}[htbp]
\centering
\caption{Two-tier DDPG Hyperparameters}
\label{tab:hdddpg_hyperparams}
\begin{tabular}{@{}llllll@{}}
\toprule
\textbf{Hyperparameter} & \textbf{Value} & \textbf{Hyperparameter} & \textbf{Value} & \textbf{Hyperparameter} & \textbf{Value} \\ \midrule
Replay Buffer & 100k & Batch Size & 256 & Learning Rate (LR) & 0.00017 \\
Discount ($\gamma$) & 0.99 & Soft Update ($\tau$) & 0.005 & Gradient Clipping & 0.8 \\
Makespan Target & 1.3s & Makespan Weight & 0.98 & Noise Decay & 0.995 \\ \bottomrule
\end{tabular}
\end{table}
\vspace{-0.3cm}
\subsection{Experimental Results}
\label{sec:exp_results}

This section presents an empirical evaluation of the proposed Two-tier DDPG scheduler. The analysis first validates the learning process, then benchmarks the scheduler's makespan performance against the baselines from Section~\ref{sec:exp_design}. The results are interpreted in the context of the core challenges motivating this work: the NP-hard complexity of the problem and the limitations of static or non-hierarchical approaches.

\begin{figure}[htbp]
    \centering
    \includegraphics[width=0.9\textwidth]{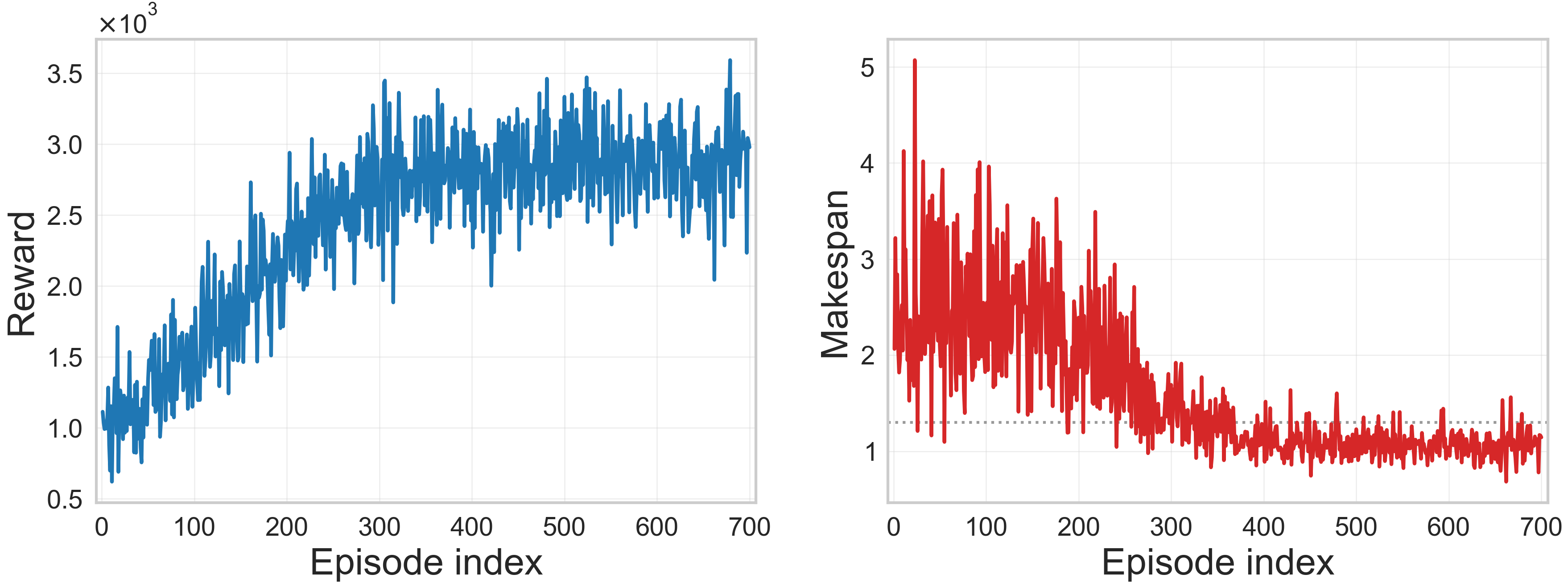}
    \caption{Training performance of the Two-tier DDPG scheduler over 700 episodes.}
    \label{fig:training_curve}
\end{figure}

The training process efficacy is depicted in \cref{fig:training_curve}. The episodic reward demonstrates stable convergence after approximately 400 episodes, mirrored by a corresponding decrease in the average makespan. This inverse correlation confirms that the reward function (\cref{eq:rl_reward_detailed}) effectively guides the agent toward the primary optimization objective (\cref{eq:obj}). Furthermore, this efficient convergence validates the architectural hypothesis of the Two-tier DDPG. By decomposing the large, flat action space into a hierarchical structure, the scheduler counters the exploration inefficiencies and poor scalability inherent to non-hierarchical reinforcement learning.

\vspace{-0.3cm}
\begin{figure}[htbp]
    \centering
    \includegraphics[width=0.8\textwidth]{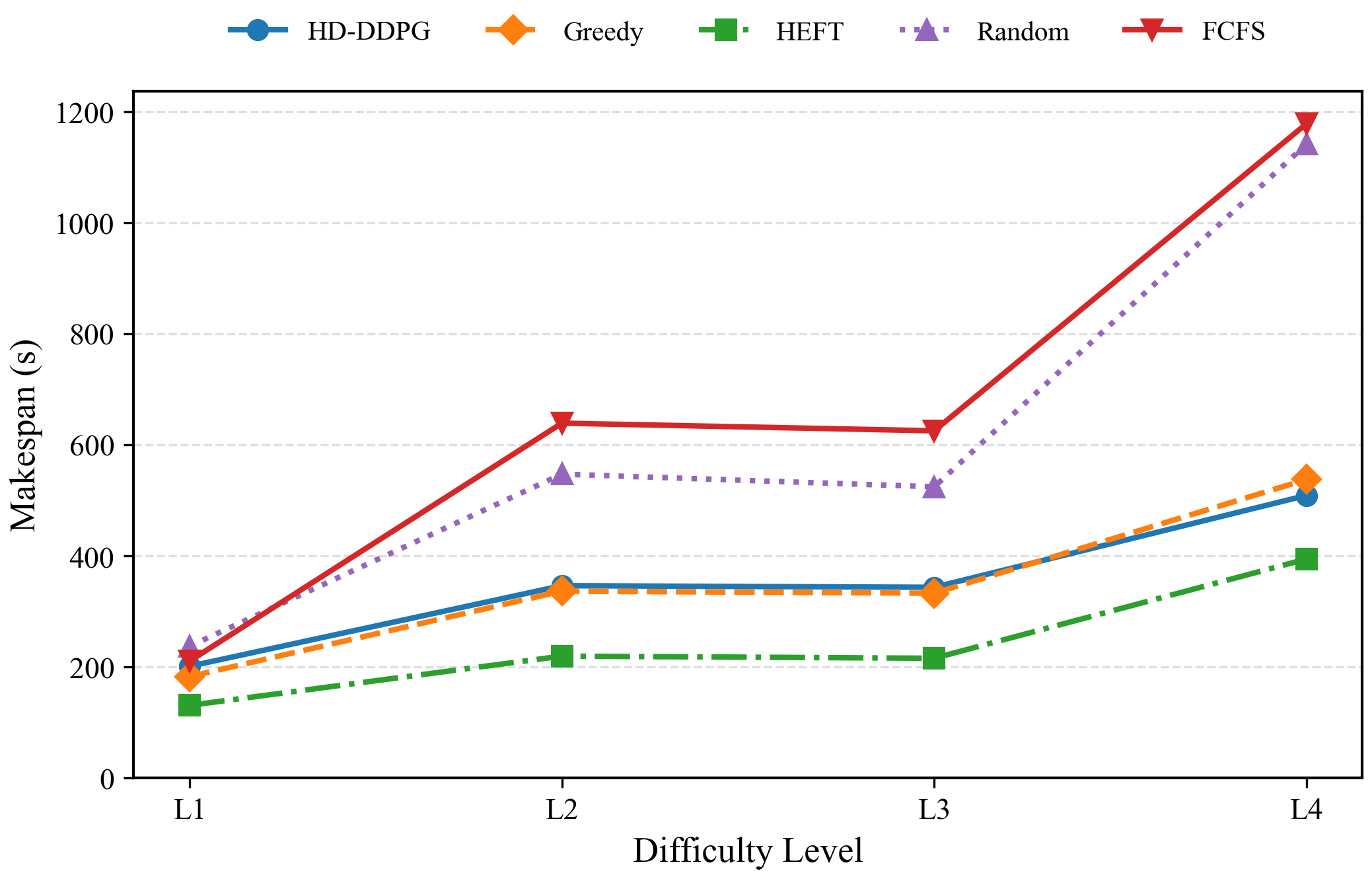}
    \caption{Total makespan comparison across four difficulty levels (L1-L4).}
    \label{fig:makespan_comp}
\end{figure}

Following training, the converged policy was benchmarked on the L1--L4 test sets, with results presented in \cref{fig:makespan_comp}. The Two-tier DDPG scheduler consistently outperforms the Random, FCFS, and Greedy baselines. Its advantage over the Greedy heuristic is particularly noteworthy as complexity increases, reaching 5.3\% at the L4 level. This divergence highlights the agent's capacity to learn superior long-term strategies for this NP-hard problem. While the Greedy approach makes locally optimal choices based on immediate cost (\cref{eq:task_cost}), the RL agent learns to account for long-term dependencies—a critical capability where the cumulative impact of decisions is more significant.The scheduler's performance relative to the static HEFT benchmark reveals another key insight. While HEFT, an offline heuristic with a global view, performs strongly, the makespan gap between it and our online Two-tier DDPG narrows from 53.8\% at L1 to 29\% at L4. This trend indicates that as workflow complexity ($|\mathcal{V}|$) increases, the adaptive decision-making of the online RL agent becomes increasingly competitive. This addresses a primary motivation for this research: developing a scheduler that can adapt to complex environments where the efficacy of pre-computed, static schedules diminishes.

\section{Conclusion}
\label{sec:conclusion}
% \subsection{Conclusion}
This paper investigated the scheduling of sequential MIoT workflows on a heterogeneous cloud-fog-edge infrastructure, a challenge characterized by its NP-hard complexity. Our core contribution is a Two-tier DDPG scheduling framework that decomposes the complex assignment task into a hierarchical process of global layer selection and local node assignment. This two-tier architecture, executed by a global controller and local controllers, aligns with the physical system's hierarchy. Our scheduler considers the computation-communication trade-off at the global level, while the decomposition of the action space enables more efficient policy learning, as validated by the stable convergence observed in the training process. Experiments results demonstrated that our scheduler consistently outperforms standard baselines like Random and FCFS, highlighting its ability to learn more effective long-term strategies compared to these approaches. Notably, while the static HEFT heuristic remains a strong benchmark, the performance gap between it and our adaptive online scheduler narrows as workflow difficulty increases. This trend underscores the potential of our learning-based framework for dynamic, large-scale MIoT environments, where the adaptability of online scheduling becomes increasingly critical.

\subsubsection{Acknowledgements} This work was supported by Hong Kong Metropolitan University, in part by the grant from the Research Grants Council of the Hong Kong Special Administrative Region, China, under project No. UGC/FDS16/E15/24, and in part by the UGC Research Matching Grant Scheme under Project No.: 2024/3003.

\bibliographystyle{splncs04}

\end{document}